\documentclass[fleqn,10pt]{wlscirep}
\usepackage[utf8]{inputenc}
\usepackage[T1]{fontenc}
\usepackage{array}
\usepackage{multirow}
\usepackage{booktabs}
\usepackage{subcaption}
\usepackage{xcolor}
\usepackage[normalem]{ulem}
\usepackage{threeparttable}

\title{Epithelium segmentation using deep learning in H\&E-stained prostate specimens with immunohistochemistry as reference standard}

\author[1,*]{Wouter Bulten}
\author[1]{P\'eter B\'andi}
\author[2]{Jeffrey Hoven}
\author[2]{Rob van de Loo}
\author[3]{Johannes Lotz}
\author[3]{Nick Weiss}
\author[1]{Jeroen van der Laak}
\author[4]{Bram van Ginneken}
\author[2]{Christina Hulsbergen-van de Kaa}
\author[1]{Geert Litjens}
\affil[1]{Radboud University Medical Center, Diagnostic Image Analysis Group and the Department of Pathology, 6500HB, Nijmegen, The Netherlands}
\affil[2]{Radboud University Medical Center, Department of Pathology, 6500HB, Nijmegen, The Netherlands}
\affil[3]{Fraunhofer MEVIS, 23562, L\"ubeck, Germany}
\affil[4]{Radboud University Medical Center, Diagnostic Image Analysis Group and the Department of Radiology and Nuclear Medicine, 6500HB, Nijmegen, The Netherlands}
\affil[*]{wouter.bulten@radboudumc.nl}

\begin{abstract}
Given the importance of gland morphology in grading prostate cancer (PCa), automatically differentiating between epithelium and other tissues is an important prerequisite for the development of automated methods for detecting PCa. 
We propose a new deep learning method to segment epithelial tissue in digitised hematoxylin and eosin (H\&E) stained prostatectomy slides using immunohistochemistry (IHC) as reference standard.
We used IHC to create a precise and objective ground truth compared to manual outlining on H\&E slides, especially in areas with high-grade PCa. 102 tissue sections were stained with H\&E and subsequently restained with P63 and CK8/18 IHC markers to highlight epithelial structures. Afterwards each pair was co-registered.
First, we trained a U-Net to segment epithelial structures in IHC using a subset of the IHC slides that were preprocessed with color deconvolution. Second, this network was applied to the remaining slides to create the reference standard used to train a second U-Net on H\&E.
Our system accurately segmented both intact glands and individual tumour epithelial cells. The generalisation capacity of our system is shown using an independent external dataset from a different centre.
We envision this segmentation as the first part of a fully automated prostate cancer grading pipeline.
\end{abstract}
\begin{document}

\flushbottom
\maketitle

\thispagestyle{empty}

\section*{Introduction}

With 1.1 million new diagnoses every year, prostate cancer (PCa) is the  most common cancer in men in developed countries \cite{Torre2015}. PCa develops from genetically damaged glandular epithelium, resulting in altered cellular proliferation patterns. In the case of high-grade tumours, the glandular structure is eventually lost and strands of (individual) cells can be observed instead \cite{Fine2012}. 

The histological grade in PCa is formally defined in the Gleason grading system \cite{Epstein2010}, and is a powerful prognostic marker. It is determined by pathologists on hematoxylin and eosin (H\&E) stained tissue specimens. The grade is based on the architectural growth patterns of the tumour which are assigned a number between 1 and 5, with increasing numbers corresponding to a decrease in histological differentiation, and, typically, worse prognosis \cite{Epstein2005}. 

The identification and grading of prostate cancer can be time consuming and tedious for pathologists, as all individual cancer foci within a surgical specimen or biopsy have to be analysed. This is compounded by the fact that prostate cancer is generally a multi-focal disease and that surgical specimens can consists of anywhere between 8 - 15 sections. Although nowadays, thanks to the advent of whole-slide scanning systems, pathologists can perform their diagnoses on a computer screen instead of using a microscope, this has not directly helped them to perform more efficient or accurate diagnostics. However, computer-aided diagnostic tools based on deep learning and convolutional neural networks have shown promise in improving the accuracy and efficiency of histopathological diagnosis \cite{Litj16c}.

Deep learning methods that try to detect or grade cancer from scanned tissue slides are typically trained using a set of annotated regions as the reference standard. As these algorithms learn from training data, the quality of the output is directly linked to the quality of the training samples. Ideally, training samples for detecting and grading PCa consist of individually outlined glands. However, outlining PCa requires extensive expert knowledge due to the large differences between and within Gleason grades. In addition, annotating individual cells of high grade PCa is practically infeasible due to the mixture of glandular, stromal and inflammatory components. Therefore, tumor annotations made by pathologists are often coarse and contain large amounts of non-relevant tissue which adds noise to the reference standard and, subsequently, limits the potential of deep learning methods.

We propose a method to automatically improve the detail of PCa annotations by pathologists by dividing digitised tissue into relevant and non-relevant tissue on a pixel-by-pixel basis, in this case epithelial versus other tissues. Such a system can help improve the detail of coarse cancer or grade annotations, but can also be useful by itself in highlighting areas containing epithelial cells as regions of interest for pathologists.

\begin{figure*}[!t]
\centering
\includegraphics[width=\textwidth]{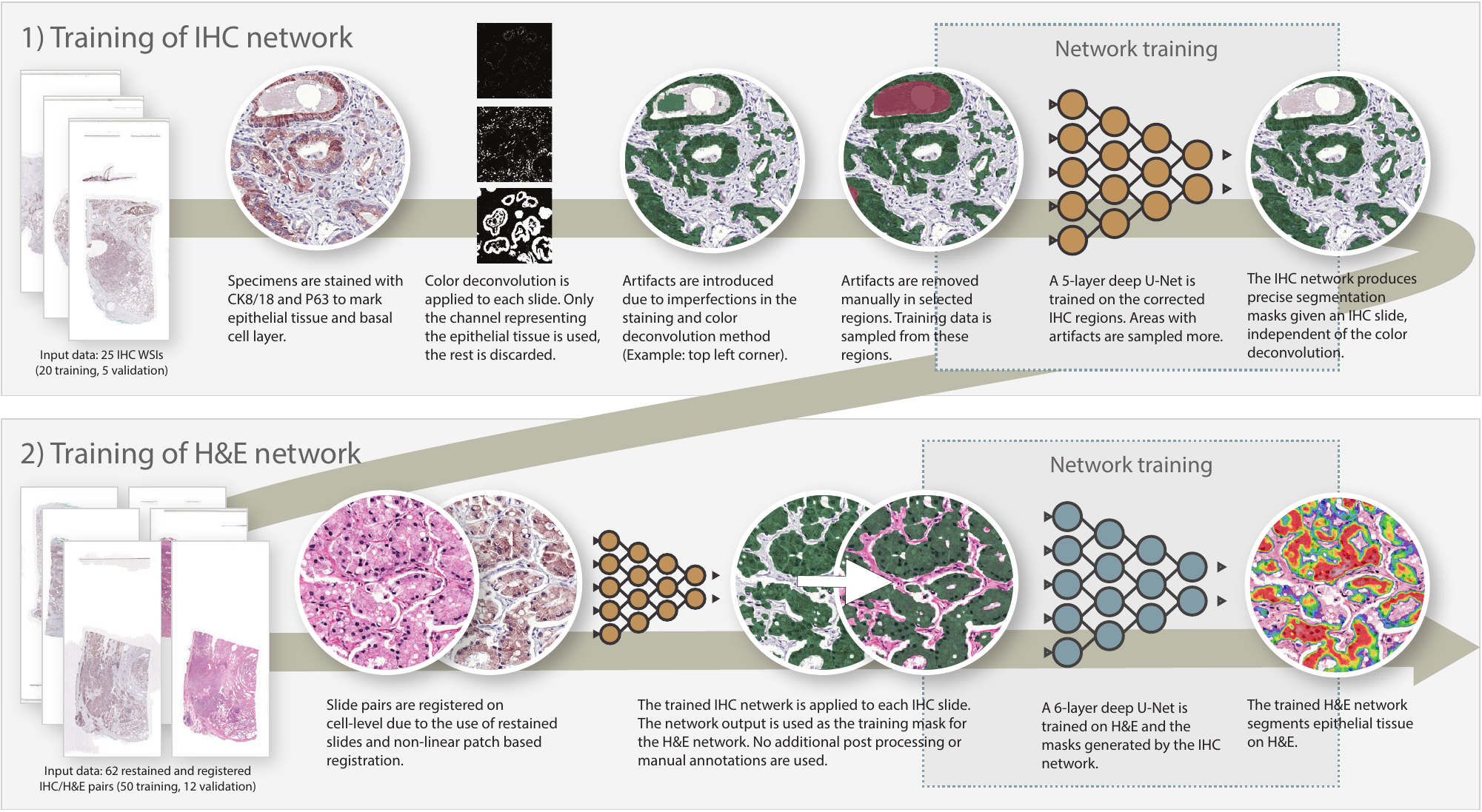}
\caption{Overview of methodology. We first train a network (1) on a subset of our IHC training data. The segmentations produced by this first network are then transferred to H\&E and used to train the final network (2).}
\label{fig:alg}
\end{figure*}

To train our system, we employed a novel two-step approach (Figure \ref{fig:alg}). First, we trained a convolutional network to segment epithelium in immunohistochemically (IHC) stained tissue sections applying an epithelial marker. By applying color deconvolution and subsequent recognition of positively stained pixels, we were able to have ample training data while obviating the cumbersome and imprecise process of manually annotating epithelial regions \cite{Ruif01, Geijs2018}. Registration was used to map the network's output to the H\&E version of the specimens which were subsequently used as training input for our final model. Our automated segmentation is not only useful as a tool for pathologists, we particularly envision this segmentation as being the first part of a fully automated prostate cancer detection and grading pipeline.

\section*{Related work}
\label{sec:relwork}

Existing research on segmenting epithelial tissue has shown promise in PCa specimens. Gertych et al.\cite{Gert15} used a support vector machine to distinguish between stroma and epithelial glands and applied this to a dataset of 20 patients containing specimens of Gleason grade 3 and 4. Hand crafted features, based on intensity and spatial relationship of pixels, were derived from H\&E specimens that had been preprocessed using color deconvolution. Naik et al.\cite{Naik2007} employ Bayesian classifiers to segment glands, relying on the presence of lumen in the glands. The segmentation was applied to Gleason grade 3 and 4, and benign tissue samples; not on the less common but more aggressive pattern 5. Gleason grade 5 can express in the form of single-cell strands or nests, or solid sheets (with or without central necrosis) of malignant cells with no or minimal lumen formation; obviously, this could hinder a segmentation method that relies on the presence of lumina. Singh et al.\cite{Singh2017} employed a multi-step approach  based on logistic regression to segment epithelium, distinguishing between glands, lumen, peri-acinar retraction clefting and stroma. Both Gertych et al.\cite{Gert15} and Naik et al.\cite{Naik2007} used the segmentation results as a first step towards automated Gleason grading.

Advances in deep learning have resulted in new methods for performing segmentation. Deep learning methods generally outperform hand crafted features on segmentation tasks in digital pathology, for example on H\&E and IHC stained breast and colon tissue specimens\cite{Xu16}. On the dataset from Gertych et al.\cite{Gert15}, Li et al. \cite{Li2017} show a clear performance increase when using deep learning models to segment PCa in comparison to classical machine learning methods. Deep learning methods also show good performances on segmenting glands, for example in colorectal tissue\cite{VanEycke2018}.

Previously, we performed a pilot study on epithelium segmentation comparing U-Net versus regular fully convolutional networks using 30 radical prostatectomy slides and a small, manually annotated, test set\cite{Bult18}. We achieved the best segmentation performance using a 4-layer-deep U-Net, but found that the performance of our network capped due to errors in the reference standard. Moreover, a low number of samples, in particular few high grade PCa specimens, limits the applicability to daily practice.

Most of the existing studies on epithelium segmentation in prostate suffer from small datasets or focus on a subset of the occurring grades. In this paper we did not exclude any Gleason grades or gland morphology.

\begin{figure*}[!t]
\centering
\includegraphics[width=\textwidth]{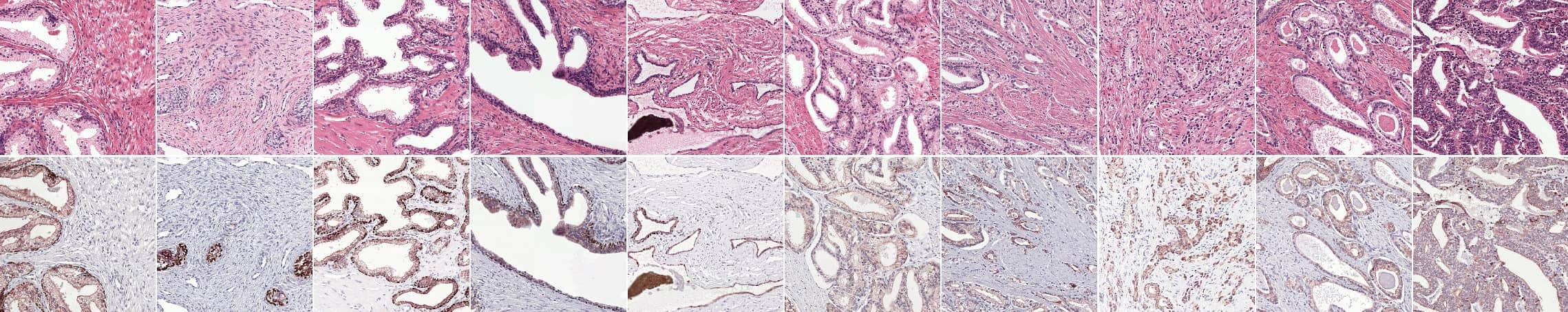}
\caption{Dataset examples (first row H\&E, IHC second). Our restaining procedure (instead of using consecutive slides) results in perfectly matching slides. The first five columns show benign epithelium, the last five show various grades of PCa. In the bottom row, all epithelial tissue is marked in brown, the basal cell layer in dark red (only present in the benign examples). Between cases the intensity of the stain can differ substantially.}
\label{fig_sim}
\end{figure*}

\begin{figure*}[t]
\centering
\begin{minipage}[t]{0.48\textwidth}
\centering
\includegraphics[width=1\columnwidth]{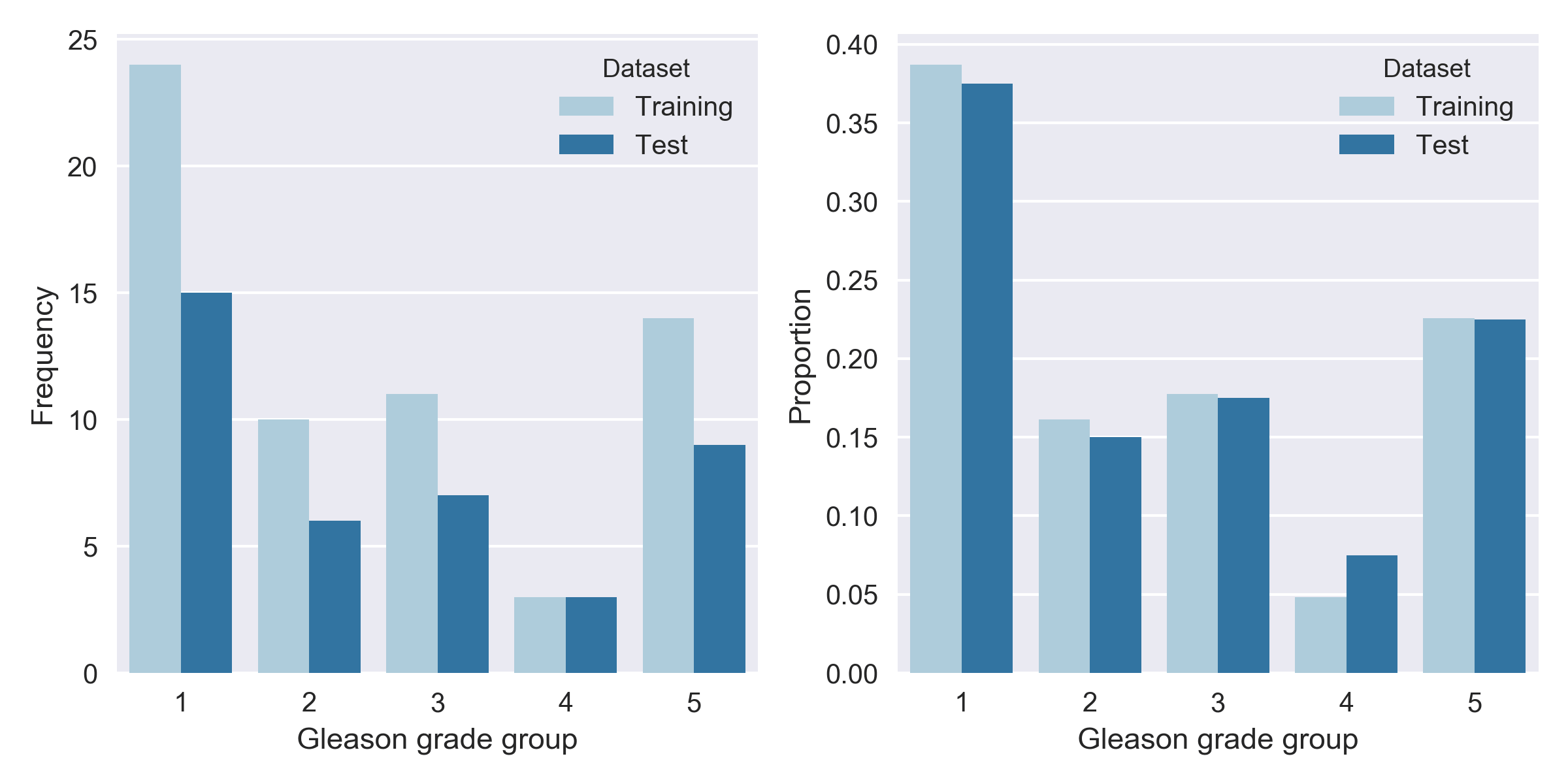}
\caption{Distribution of Gleason grade groups for each case in our dataset as reported in the original pathologist's report (N=102).}
\label{fig:data_dist}
\end{minipage}\hspace{0.04\textwidth}%
\begin{minipage}[t]{0.48\textwidth}
\centering
\includegraphics[width=1\columnwidth]{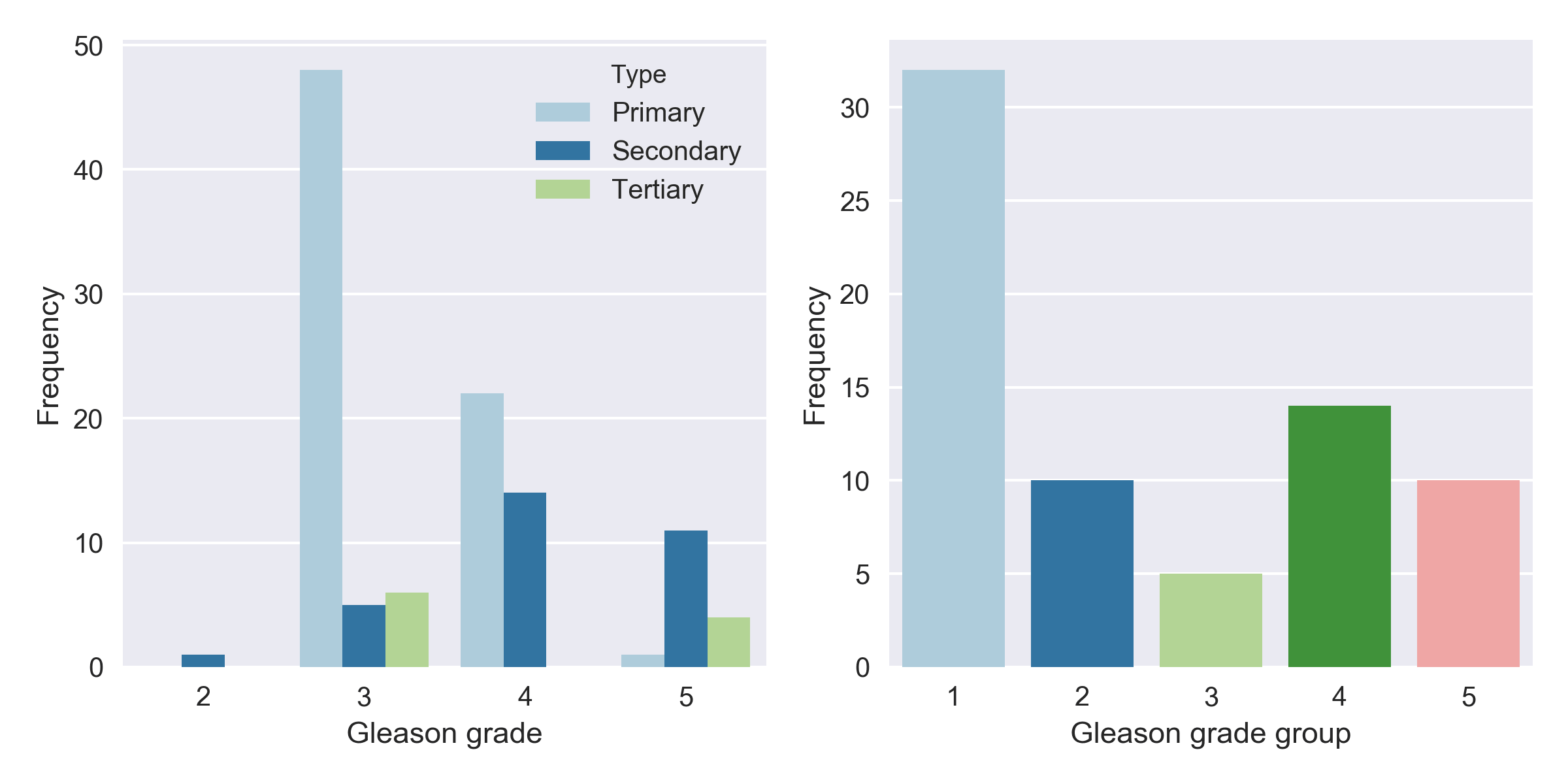}
\caption{Gleason grades of tumor regions in the hold-out test set (N=71). Showing individual occurences (left, 1-3 per region) and grade groups on region level (right).}
\label{fig:data_dist_testset}
\end{minipage}
\end{figure*}

\section*{Materials}
\label{sec:materials}

We selected a cohort of 102 patients who underwent a radical prostatectomy at the Radboud university medical center (Radboudumc) between 2006 and 2011 (IRB number 2016-2275). Patients who received adjuvant therapy before surgery were excluded. From each prostatectomy, we selected one formalin fixated paraffin embedded tissue block based on the Gleason grades reported in the original pathologist's report. Based on the reported grades, we determined the Gleason grade group \cite{Epst16} for each block (Table \ref{tab:cases}). As a tissue block can contain multiple grades we also reported the individual occurrences of each grade. Of all tissue blocks, 24\% contained a region with grade 2, 69\% with grade 3, 63\% with grade 4 and 33\% with grade 5. Due to selective oversampling, the incidence of high grade tumours (grades 4 and 5) is relatively higher than in clinical practice. This oversampling allows us to explicitly investigate the performance of deep learning based epithelium segmentation algorithms on high-grade PCa, in which such segmentation is most challenging.

\begin{table}[t]
\centering
\caption{Overview of case grading from original pathologist's report on section level (using grade group) and on individual grade. Note that multiple grades can occur within a single slide.}
\label{tab:cases}
\begin{tabular}{@{}ll*{9}{>{\centering\arraybackslash}c}@{}}
\toprule

\textbf{}    & \textbf{} & \multicolumn{5}{c}{\textbf{\begin{tabular}[c]{@{}c@{}}Grade group\\ (Section)\end{tabular}}} & \multicolumn{4}{c}{\textbf{\begin{tabular}[c]{@{}c@{}}Grade\\ (Individual)\end{tabular}}} \\ \cmidrule(lr){3-7} \cmidrule(lr){8-11} 
\textbf{Set} & \textbf{\# slides}        & \textbf{1}         & \textbf{2}        & \textbf{3}        & \textbf{4}        & \textbf{5}        & \textbf{2}                 & \textbf{3}                & \textbf{4}                & \textbf{5}                \\ \midrule

Train set            & 62                        & 24               & 10               & 11               & 3              & 14              & 12                      & 44                      & 40                      & 22                      \\
Test set             & 40                        & 15               & 6                & 7                & 3              & 9               & 12                      & 26                      & 24                      & 12                      \\ \midrule
Total                & 102                       & 39               & 16               & 18               & 6              & 23              & 24                      & 70                      & 64                      & 34                      \\ \bottomrule
\end{tabular}
\end{table}

From each block a new section was cut, stained with H\&E and scanned using a \textit{3DHistech Pannoramic Flash II 250 scanner}. After scanning, the tissue was destained, restained using immunohistochemistry, and scanned again. All slides were scanned at 20x magnification (pixel resolution $0.24\, \mu m$). 

We used two markers for the immunohistochemistry: CK8/18 (using DAB) to mark all glandular epithelial tissue (benign and malignant), and P63 (using NovaRED) for the basal cell layer, which is normally present in benign glands but not in malignant glands. This staining procedure results in a slide where all relevant tissue is highlighted, providing us with a clear ground truth (see Figure \ref{fig_sim} for examples). Staining the basal cell layer using a different colour makes it easier to spot tumour regions in IHC and can facilitate grading of the tissue on H\&E. Restaining, instead of making consecutive slides, results in an H\&E and IHC whole-slide image (WSI) pair for each patient that contains the same tissue. Although the slide pairs were made from the same glass slide, minor alignment errors and tissue deformations were still present due to the restaining procedure.

The 102 scanned slide pairs were split into two sets: a training set (62) and a test set (40). The slides were distributed over the sets at random while stratifying for Gleason grade group (Figure \ref{fig:data_dist}). The test set was used as a hold-out set and not used during training or model optimisation.

\subsection*{Hold-out test set}

For each IHC slide in the test set, a trained non-expert divided each WSI in four sections: two containing tumor and two containing only benign epithelium. From each of these four regions, we extracted an area of $2500\times2500$ pixels randomly at 10x magnification. If there was either no tumor or benign region available, an additional region from the other category was selected. This method resulted in 160 regions. 

The tumor regions were individually graded by an experienced pathologist (C.H.-v.d.K.) with subspecialty uropathology, without using the original patient's record. We recorded the primary, secondary and tertiary (if present) grade for each region (Figure \ref{fig:data_dist_testset}). The reported grades were not necessarily identical to those from the patient records; the selected regions contained a subset of the slide and were extracted from a newly cut section. The Gleason grade group was based on the ISUP scoring system for biopsies (most prevalent plus highest grade).

\subsection*{External test set}

Gertych et al.\cite{Gert15} made their dataset available to use for external validation. This set consists of 224 $1500 \times 1500$ pixels tiles sampled from 20 digitised WSIs (pixel resolution $0.5\, \mu m$) of H\&E prostatectomy specimens containing Gleason grades 3 and 4. The tiles were already annotated by two pathologists and each pixel labelled as stroma, benign epithelium, Gleason 3 or Gleason 4. Glands were annotated as a whole, including the lumen. We combined the annotations of benign epithelium and the two PCa grades into a single epithelium class.

\section*{Methods}
\label{sec:methods}

We took a two-step approach to train a system for segmentation of epithelial tissue on H\&E histopathology. First, we circumvented the challenge of manually annotating tissue by generating precise training data using immunohistochemistry and training a network on IHC. Then we transferred the output of the first network to H\&E and trained the final segmentation network. Our networks were built using \textit{Keras} \cite{chollet2015keras} and \textit{Tensorflow} \cite{tensorflow2015}.

\subsection*{Slide preparation}

We applied a pre-trained tissue-background segmentation network \cite{Band17} to all slides in order to exclude areas not containing tissue from further analysis. Next, color deconvolution was applied to all IHC WSIs in our training set \cite{Ruif01, Geijs2018}. The resulting P63/CK8-18 channel was then converted to a binary mask by thresholding. Small errors were removed automatically using binary closing and opening. The resulting masks were not perfect due to imperfections and intensity changes in the stain, scanning artefacts and non-specific staining; e.g. corpora amylacea and debris inside the glands are regularly stained brown and are therefore present in the deconvolution mask (Figure \ref{fig:stain_artifact}a,c).

For the hold-out test set, three trained non-experts reviewed the sampled test regions and manually updated the color deconvolution mask, removing any artefacts or updating incorrectly labeled tissue.

\subsection*{Training a CNN on IHC}
Due to time-constraints, it was unfeasible to manually correct all individual color deconvolution masks to be used for training. Instead, we trained a deep convolutional network to perform the mapping from a P63/CK8-18 slide to a binary epithelium mask. We selected 25 slides from our training set to train this first network (20 for training, 5 for validation). On each slide we outlined a tissue region covering roughly 50\% of the WSI after which three trained non-experts corrected the color deconvolution masks by hand. A total of 3493 annotations were made by the annotators on these 25 slides, an average of 140 annotations per slide. In terms of surface area, $2.3\%$ of the tissue was given a different label by the annotators. On average, the annotators took 45 to 60 minutes to correct a slide.

We trained a five-level-deep U-Net \cite{Ronn15} on the selected regions to segment epithelial tissue in IHC slides. We followed the original U-Net model architecture, but added additional skip connections within each layer block, and used up-sampling operations in the expansion path. The network was trained using randomly sampled patches with a size of $512\times512$ (pixel resolution $0.48\, \mu m$) and a batch size of 1. Regions with annotated artefacts and corpora amylacea were oversampled to lower the number of false positives. Adam optimisation was used with $\beta_1$ and $\beta_2$ set to $0.99$, and a learning rate of $0.0005$. The learning rate was halved after every 5 consecutive epochs without improvement on the validation set.

During training, we applied data augmentation to prevent overfitting and to improve the model's generalisation. The following augmentations were used: flipping, rotation, additive Gaussian noise, Gaussian blurring and changes in saturation, contrast and brightness. After training, the model was applied to all IHC WSIs in our training set. A binary mask was created from each slide using the argmax of the network output. We focused explicitly on colour augmentations to overcome the large stain differences between the IHC slides.

For comparison, a second U-Net was trained on the non-corrected colour deconvolution masks directly, without using any of the manual corrections. All hyperparameters and network structure were kept the same as in the original experiment to create a fair comparison.

\subsection*{Registration}

The H\&E slides were registered to the IHC slides using a nonlinear image registration method based on a method described previously \cite{lotz_patch-based_2016}. Since both slide images showed the same object with different stains, they were already approximately aligned. However, additional nonlinear deformations are caused by the chemical treatment during restaining and/or the slide scanning procedure and needed to be compensated for. Since different stains are used in both images, the colours of spatially corresponding structures do not match (Figure \ref{fig_sim}). We use the Normalised Gradient Fields (NGF) distance \cite{haber_intensity_2007}, that measures the alignment of image gradients, to account for the multi-modality of the registration problem. 

The registration pipeline consisted of: conversion of RGB images to gray-scale $\rightarrow$ parametric (affine) registration $\rightarrow$ nonparametric registration (NGF distance measure \cite{haber_intensity_2007}, curvature regulariser \cite{fischer_curvature_2003}) $\rightarrow$ patch-based registration (NGF, curvature). The method to merge the patches has been extended as follows: Instead of averaging the deformation patches, an optimisation problem is solved that balances data-fit and global deformation regularisation in the overlap region.

\subsection*{Training a CNN on H\&E}

The training masks generated by the IHC network matched the H\&E slides as a result of the registration step; 50 were used for training and 12 for validation. We found that increasing the depth of the U-Net lowered the number of misclassified corpora amylacea on H\&E. Therefore, for the H\&E segmentation we trained a six-level-deep U-Net in comparison to the five-level-deep IHC network. To limit the parameter count caused by the added level we lowered the amount of filters for each level. The same extensions as used in the U-net for the IHC stained images were applied. The network was trained using patches with a size of $1024\times1024$ (pixel resolution $0.48\, \mu m$) and a batch size of 1. Adam optimisation was used with $\beta_1$ and $\beta_2$ set to $0.99$, and a learning rate of $0.0005$. The learning rate was halved after every 10 consecutive epochs without improvement on the validation set. The following data augmentations were used: random scaling, flipping, rotation, additive Gaussian noise, Gaussian blurring and changes in saturation, contrast, brightness and Haematoxylin-Eosin colour space.

Only the binary segmentation masks generated by the IHC network were available for training. We did not correct the masks manually. This meant that the sampling technique used for training the IHC network could not be applied to the H\&E network. Instead we sampled uniformly over the classes. To force the network to learn small areas of epithelium, e.g. in cases of Gleason 5, we weighted the loss of each pixel based on the class occurrence within a patch. As a result, even patches with only small individual tumor cells were picked up by the network due to a higher loss contribution. 

To test the merit of the IHC network as input for our network, we also trained a U-Net on the raw color deconvolution masks. All hyperparameters and network structure were kept the same in both experiments.

\subsection*{Evaluation}
    
 The trained H\&E network was applied to all WSIs of our hold-out set and evaluated within the randomly selected regions. No further post-processing was performed.

The annotations of the external set were coarse and on gland-level (i.e. including the lumina) and did not match the output of our network. In accordance with the method used in the original paper, we removed the background from the color-normalised images of the external test set \cite{Gert15}. Lumina (consisting of pixels which are classified as background pixels) were not used in computing the scores.  We then fed the images to our trained H\&E network. We did not optimise our network on this external set. As such, the results on the external test set can be considered a true estimate of the generalisation capacity of our H\&E network.

\begin{figure*}[!t]
\centering
\includegraphics[width=\textwidth]{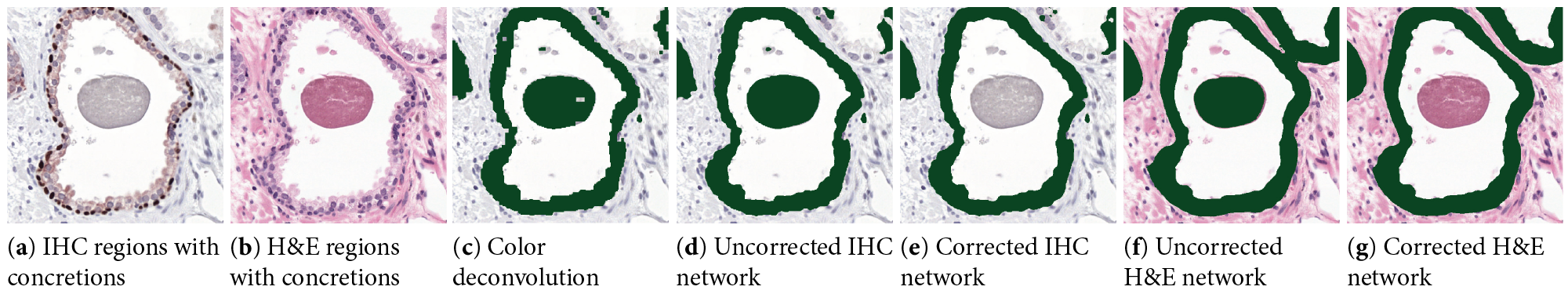}
\caption{Effect of stain artefacts on network predictions. In some cases non-epithelial tissue is stained, e.g. structures inside the gland (a, corresponding H\&E version shown in b). These artefacts are also picked up by the color deconvolution algorithm (c). Due to a high frequency of these artefacts, training a network on this uncorrected data results in a trained network that has a high occurrence of false positives in its predictions (d). Training a network on manually corrected data instead, results in a better segmentation (e). These errors transfer to the training of the H\&E network. A network trained on the raw color deconvolution masks makes more mistakes in these artefact regions (f) than a network trained on the output of the corrected IHC network (g).}
\label{fig:stain_artifact}
\end{figure*}

\section*{Results}
\label{sec:results}

We evaluated both the IHC and H\&E networks on the regions from, respectively, the IHC and the H\&E WSIs from the hold-out test set. The network output was compared with the ground truth: color deconvolution masks generated from the IHC slides with manual corrections. We report pixel-based accuracy, F1-score and Jaccard index using epithelium as the positive label (Table \ref{tab:results}).

\begin{figure}[t]
\centering
\includegraphics[width=\textwidth]{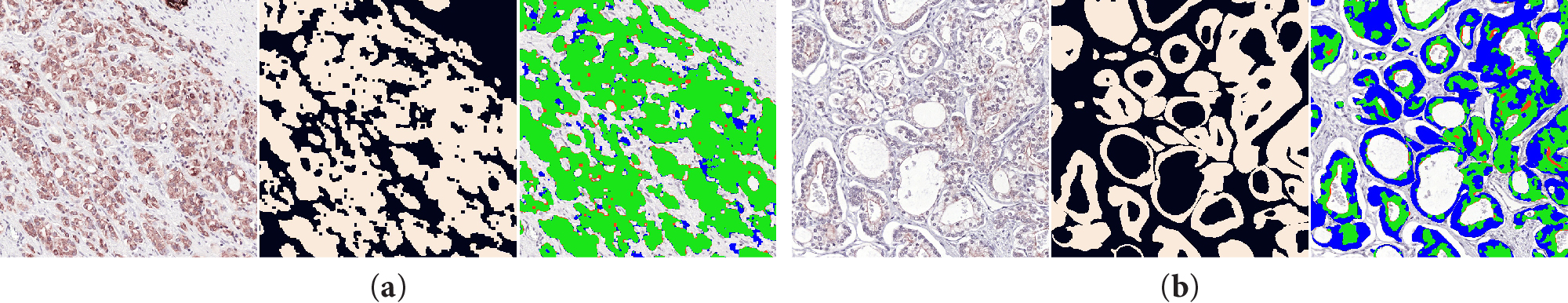}
\caption{Zoomed-in examples ($1000\times1000$ crop) of the hold-out test set: IHC version (left), ground truth (middle) and segmentation of the IHC network (right). Green pixels show true positive, red false positive and blue false negative. The first example (a) shows an almost perfect segmentation. In regions where the stain is light or absent the performance degrades (b).}
\label{fig:testset_ihc}
\end{figure}

\subsection*{Segmentation performance on IHC}

The IHC network achieved an overall F1 score of 0.915. Given a minimum F1 score of 0.352 and a maximum of 0.980, the range of scores was high. Some regions of our test set suffered from an overall low stain quality or contained areas where the epithelium reacted less to the stain. We observed that a lower stain intensity resulted in a lower performance (Figure \ref{fig:testset_ihc}). As the H\&E network was trained on the output of the IHC network we considered the IHC performance as an upper bound for the performance of the H\&E network.

Using the corrected color deconvolution masks as training data resulted in an F1 score increase of 0.909 to 0.915 on our test set (Table \ref{tab:result_corrections}). The network that was trained on the uncorrected data makes more mistakes in regions with stained non-epithelial tissue, e.g. corpora amylacea and other concretions inside the glands (Figure \ref{fig:stain_artifact}d).

\subsection*{Segmentation performance on H\&E}

\begin{figure*}[!t]
\centering
\includegraphics[width=\textwidth]{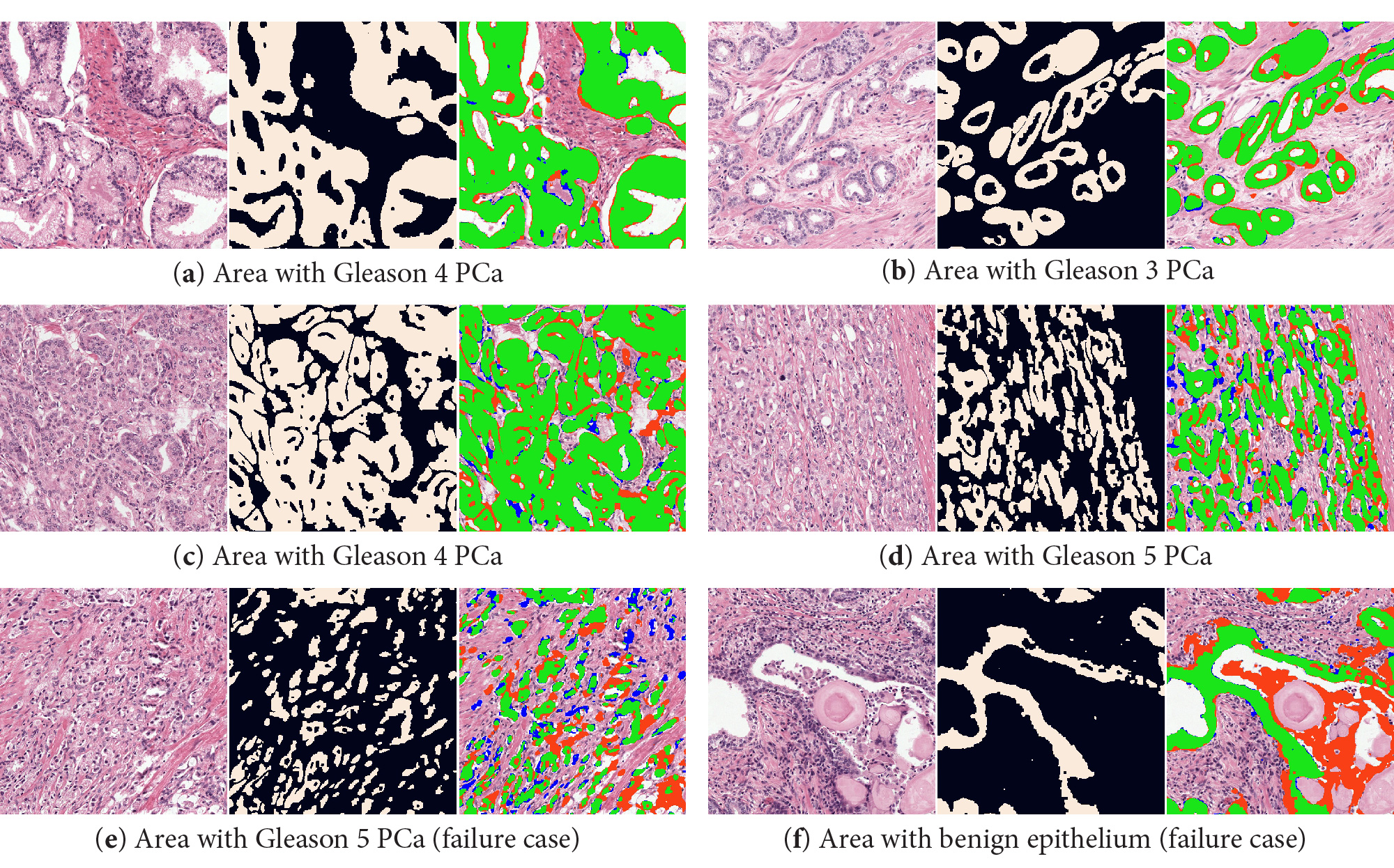}
\caption{Zoomed-in example regions ($1000\times1000$ crop) from the hold-out test set with H\&E (left), ground truth (middle) and network segmentation (right). Green pixels show true positive, red false positive and blue false negative. The top two rows displays two cases (a, b, c, d) of PCa where the network segments the epithelial tissue almost perfectly. In the bottom row two failure cases are shown: a case of high grade PCa (e) and a benign region (f) where debris inside the gland is segmented.}
\label{fig:testset_overlay}
\end{figure*}

The H\&E network achieved an overall F1 score of 0.893. The score on benign tissue (F1 0.907) was slightly higher than on tumorous areas (F1 0.876). A decline in performance was observed in regions with higher Gleason grades. Regions with Gleason grade group 5 had an F1 score of 0.819. Several regions are displayed in Figure \ref{fig:testset_overlay}.

The score of the H\&E network was comparable to that of the IHC network, showing that, given this training data, the network achieved an almost optimal performance. Even more, the minimal performance of the H\&E network was higher than the minimum of the IHC network (0.661 versus 0.352). Outliers that were present in the results of the IHC network were not present in the results of the H\&E network.

Using the IHC network to generate training data, as opposed to the raw color deconvolution masks, resulted in an improved F1 score of 0.893 versus 0.878 for the uncorrected network (Table \ref{tab:result_corrections}). Comparable to the IHC network, the uncorrected H\&E network makes more mistakes in areas that are incorrectly targeted by the stain (Figure \ref{fig:stain_artifact}f).

\begin{table}[]
\centering
\caption{Segmentation results on the hold-out test set.}
\label{tab:results}
\begin{tabular}{@{}lllll@{}}
\toprule
\textbf{Regions} & \textbf{N} & \textbf{F1 score}                & \textbf{Accuracy} & \textbf{Jaccard}\\
\textbf{}        & \textbf{}  & \textbf{mean (min, max)}& \textbf{}         \\ \toprule
\textbf{IHC network} &&& \\ \midrule
All regions      & 160        & $0.915 \pm .09$ (0.352, 0.980)        & 0.952 & 0.854  \\ \midrule
Benign           & 89         & $0.944 \pm .04$ (0.712, 0.980)       & 0.980 &  0.897 \\
Cancer           & 71         & $0.879 \pm .11$ (0.352, 0.974)       & 0.917 & 0.799 \\ \midrule
\textbf{H\&E network} &&& \\ \midrule
All regions      & 160        &  $0.893 \pm .05$ (0.661, 0.959)  & 0.940 & 0.811 \\ \midrule
Benign           & 89         & $0.907 \pm .04$ (0.780, 0.957)& 0.966& 0.832\\
Cancer           & 71         & $0.876 \pm .05$ (0.661, 0.959)       & 0.907 & 0.784 \\ \midrule
Grade group 1            & 32         & $0.884 \pm .03$ (0.808, 0.938)      &  0.921 & 0.793 \\
Grade group 2            & 10         & $0.885 \pm .03$ (0.854, 0.927)       &  0.894 & 0.794\\
Grade group 3            & 5          &  $0.893 \pm .03$ (0.833, 0.921)
       & 0.912 & 0.809 \\
Grade group 4            & 14         &  $0.889 \pm .06$ (0.728, 0.959)
      & 0.907 & 0.806 \\
Grade group 5            & 10         & $0.819 \pm .07$ (0.661, 0.914)
    &0.874 &  0.699\\ \bottomrule
\end{tabular}
\end{table}

\begin{table}[]
\centering
\caption{Comparison of segmentation performance of networks trained on the raw color deconvolution masks or using corrected training data.}
\label{tab:result_corrections}
\begin{tabular}{@{}llll@{}}
\toprule
\textbf{Training data}  & \textbf{F1 score}                & \textbf{Accuracy} & \textbf{Jaccard}\\
\textbf{}   & \textbf{mean (min, max)}& \textbf{}         \\ \toprule
\textbf{IHC network} && \\ \midrule
Color deconvolution        & $0.909 \pm .10$ (0.312, 0.983) & 0.951 & 0.844\\
Color deconvolution + corrections        & $0.915 \pm .09$  (0.352, 0.980) & 0.952 & 0.854\\ \midrule
\textbf{H\&E network} && \\ \midrule
Color deconvolution        & $0.878 \pm .06$ (0.650, 0.954) & 0.933 & 0.787\\
IHC network predictions        & $0.893 \pm .05$ (0.661, 0.959) & 0.940 & 0.811\\ \bottomrule
\end{tabular}
\end{table}

\begin{table}[]
\centering
\caption{Comparison of results on the external test set. Note that our method has not been trained on this external set while the other methods have been trained using cross validation.}
\label{tab:extset}
\begin{threeparttable}
\begin{tabular}{@{}l@{}llll@{}}
\toprule
\textbf{Network}     & \textbf{Evaluation}                   & \textbf{Accuracy}         & \textbf{F1}               & \textbf{Jaccard}          \\ \midrule
Gertych et al.  \cite{Gert15} & Cross-validation &-                & -                & $0.595 \pm .15$ \\
Li et al. \cite{Li2017} & Cross-validation &-                & -                &$0.737$\textsuperscript{*} \\
Our method                 & Hold-out validation & $0.866 \pm .07$&  $0.835 \pm .13$ & $0.735 \pm .16$ \\ \bottomrule
\end{tabular}
\begin{tablenotes}\footnotesize
\item[*] Li et al. reported separate scores for segmenting benign and cancerous epithelium. The score displayed here is the average of those two.
\end{tablenotes}
\end{threeparttable}
\end{table}

\begin{figure}
\centering
\includegraphics[width=\textwidth]{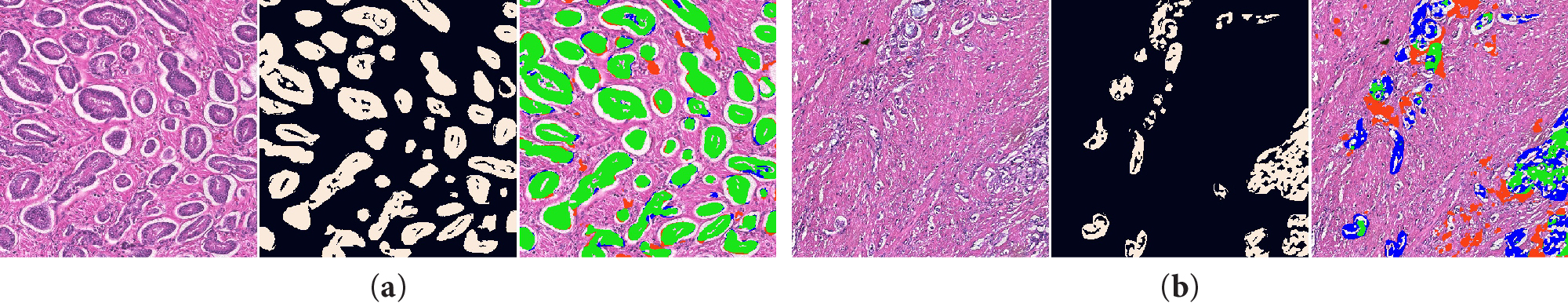}
\caption{H\&E network applied to cases from the external test set: original image (left), ground truth with background removed (middle) and segmentation of the H\&E network (right). The first example (a) shows an example of a good segmentation, the second (b) a case of undersegmentation.}
\label{fig:ext_testset}
\end{figure}

\subsection*{Segmentation performance on external dataset}

On the external set our network achieved an F1 score of 0.835 (Table \ref{tab:extset}, Figure \ref{fig:ext_testset}). This is lower than on our hold-out test set, but within expectations due to the differences in staining and image resolution. With a Jaccard score of 0.735 we achieved a higher score than the original method\cite{Gert15}, which had a Jaccard score of 0.595, and comparable to other deep learning methods that have been trained on this dataset\cite{Li2017}.

\section*{Discussion}

We developed a deep learning based system that segments epithelial tissue in H\&E-stained whole-slide prostatectomy images. Our system produces cell-level segmentations and is able to segment both intact glands as well as individual (tumor) epithelial cells. A common problem when training deep learning models for scanned histology sections is the absence of a precise ground truth. We circumvented this problem by restaining our slides with an epithelial and basal cell layer marker. Using color deconvolution and a separately trained network we were able to exhaustively annotate our complete training set with only a minimal amount of manual labour. This technique works especially well for annotating small instances of epithelium, e.g. cases of Gleason 5 PCa, that would most likely be missed by human annotators. Moreover, use of specific markers renders our ground truth less subjective compared to manually produced annotations on H\&E slides (even in inflamed or poorly differentiated areas). On an external test set we see a drastic performance improvement compared to the original method, showing the generalisation capacity of our network, even on images from an external centre. When comparing to more recent deep learning methods on this dataset, we observe that our method performs as good. Of notice is that the methods we compare against were trained on the external test dataset (in cross-validation), whereas our network has never seen this data before.

In contrast to other previous work, we assess the performance of our algorithm across all Gleason grades, including the notoriously difficulty Gleason grade 5. Although we do obtain the lowest score on this pattern (F1-score of 0.819), this score is still high especially given the poorly differentiated character of high Gleason grades, and the first benchmark on these grades. To allow others to compare their algorithms against ours we have decided to release our test data and H\&E WSIs publicly, including both the test and training slides\cite{Bult18Peso}. This dataset includes the 102 whole-slide H\&E images used in this paper, all color deconvolution masks and the manually corrected regions.

We train our IHC network on manually corrected regions which adds additional effort to the training procedure. These manual annotations result in a small increase in performance on our test set (F1 score 0.915) in comparison to training on non-corrected data (F1 score 0.909). Using the IHC network output to train the H\&E network also improves its segmentation performance (F1 score 0.893 versus 0.878). While the numerical differences are small, using the corrected data is of importance in this particular dataset to lower the number misclassifications that are caused by an aspecific stain (Figure \ref{fig:stain_artifact}) or in regions where the stain is absent. These consistent errors lower the applicability of the network in future systems. For other datasets, where stain artefacts are less prominent, training a network directly on the color deconvolution mask could be sufficient.

Our work also has some limitations. The method to establish the training labels is not perfect. The IHC network is only trained on a limited set of WSIs and is therefore not able to overcome all problems caused by stain variability and presence of scan and tissue artefacts. Especially corpora amylacea or other debris inside glands, which are often stained by the epithelial marker, are a source of errors. Epithelium glands are also missed by the network when the stain is light or absent. Subsequently, misclassified areas on the IHC slides are transferred to the training data of the H\&E network. Many of these errors are overcome by the H\&E network due to the larger size of the H\&E training set, which results in a much higher minimum performance with an F1-score of 0.661 vs. 0.352 for the IHC network.

The type of misclassifications is also influenced by the chosen magnification level. A low magnification is sufficient for segmenting intact glands, and could potentially help with lowering the number of artefacts as the network can learn high level shapes of the tissue. However, segmenting individual epithelial cells, especially in the case of high grade PCa, requires input patches with enough detail to be able to distinguish those cells from the surrounding stroma. We deliberately chose a high magnification level to improve the performance on high grade PCa. In future work it might be fruitful to investigate multi-scale approaches to tackle this issue.

We observe that the segmentation performance of our H\&E network approaches that of the IHC network, which is used to generate the training reference for the H\&E network. As a result, there is only a limited amount of improvement possible without further refining the training data. Annotating specific regions that are troublesome and retraining the IHC network on these regions could further boost the performance of the H\&E network. However, one needs to consider that for some cells it is simply impossible to assess their class using the H\&E stain alone, especially in areas with active inflammation. As such a perfect segmentation does not exist.

We see the development of an accurate epithelium segmentation network as the first part of a fully automated prostate cancer detection and grading pipeline. More specifically, the epithelium segmentation can be used to precisely outline potential cancer regions, and in combination with coarse tumor annotations result in highly detailed annotations of PCa. We intend to leverage this to develop highly accurate PCa segmentation networks in the near future.

\section*{Data availability}

The dataset generated during the current study is available in the Zenodo repository, \url{http://doi.org/10.5281/zenodo.1485967}.

\bibliography{bibliography}

\section*{Acknowledgments}

This study was financed by a grant from the Dutch Cancer Society (KWF), grant number KUN 2015-7970. The authors would like to thank Milly van den Warenburg and Nikki Wissink for their help making the manual annotations.

\section*{Author Contributions}
W.B. performed the experiments, analysed the results and wrote the manuscript. P.B. was involved in programming parts of the experimental setup. J.H. and R.v.d.L. performed data collection and annotation. J.L. and N.W. created the registration software. C.H.v.d.K. graded all cases in the test set. G.L., C.H.-v.d.K., J.v.d.L. and B.v.G. supervised the work and were involved in setting up the experimental design. All authors reviewed the manuscript and and agree with its contents.

\section*{Additional information}

\textbf{Competing interests} The authors declare no competing interests.

\end{document}